\documentclass[letterpaper, 10 pt, conference]{ieeeconf}  
\pdfoutput=1 

\IEEEoverridecommandlockouts                              

\usepackage{graphics} 
\usepackage{amsmath} 
\usepackage{verbatim}
\usepackage{float}
\usepackage[pdftex]{graphicx}
\usepackage{amsfonts}
\usepackage{caption}
\usepackage{subcaption}
\usepackage{algpseudocode}
\usepackage[linesnumbered,ruled,vlined]{algorithm2e}
\usepackage{url}
\usepackage{epstopdf}
\usepackage{romannum}

\usepackage[noadjust]{cite} 

\title{\LARGE \bf
Pose Estimation for Texture-less Shiny Objects in a Single RGB Image Using Synthetic Training Data
}

\author{Chen Chen$^{1}$, Xin Jiang$^{1}$, Weiguo Zhou$^{1}$, and Yun-Hui Liu$^{2}$
\thanks{This work was in part supported by the Shenzhen Peacock Plan Team (Grant KQTD20140630150243062) and in part supported by Shenzhen and Hong Kong Joint Innovation Project (Grant
	SGLH20161209145252406) }
\thanks{$^{1}$C. Chen, X. Jiang, W.G. Zhou are with the Department of Mechanical Engineering and Automation, Harbin Institute of Technology, Shenzhen, China {\tt\small chenchen921103@outlook.com, x.jiang@hit.edu.cn, weiguochow@gmail.com}}%
\thanks{$^{2}$Y.-H. Liu is with T Robotics Institute, The Chinese University of Hong Kong, and also visiting The State Key Laboratory of Robotics and System, Harbin Institute of Technology, Harbin, China
        {\tt\small yhliu@mae.cuhk.edu.hk}}%
}

\begin{document}

\maketitle
\thispagestyle{empty}
\pagestyle{empty}

\begin{abstract}
In the industrial domain, the pose estimation of multiple texture-less shiny parts is a valuable but challenging task. In this particular scenario, it is impractical to utilize keypoints or other texture information because most of them are not actual features of the target but the reflections of surroundings. Moreover, the similarity of color also poses a challenge in segmentation. In this article, we propose to divide the pose estimation process into three stages: object detection, features detection and pose optimization. A convolutional neural network was utilized to perform object detection. Concerning the reliability of surface texture, we leveraged the contour information for estimating pose. Since conventional contour-based methods are inapplicable to clustered metal parts due to the difficulties in segmentation, we use the dense discrete points along the metal part edges as semantic keypoints for contour detection. Afterward, we exploit both keypoint information and CAD model to calculate the 6D pose of each object in view. A typical implementation of deep learning methods not only requires a large amount of training data, but also relies on intensive human labor for labeling the datasets. Therefore, we propose an approach to generate datasets and label them automatically. Despite not using any real-world photos for training, a series of experiments showed that the algorithm built on synthetic data perform well in the real environment.
\end{abstract}

\section{INTRODUCTION}
Metal parts are essential components of many products. Enabling computers to identify and locate metal parts is a prerequisite for many industrial robotic applications, such as pick-and-place, assembly and quality control. However, the texture-less nature of metal parts poses a great challenge for many conventional pose estimation methods due to their heavy reliance on surface features. 

\begin{figure}[t]
	\centering
	\begin{subfigure}{0.237\textwidth}
		\centering
		\includegraphics[width=\textwidth]{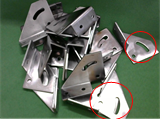}
		\caption{Differences in Color}
		\label{fig:challenge_1a}
	\end{subfigure}
	\hfill
	\begin{subfigure}{0.237\textwidth}
		\centering
		\includegraphics[width=\textwidth]{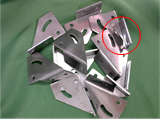}
		\caption{Reflections and Occlusions}
		\label{fig:challenge_1b}
	\end{subfigure}
	\caption{Challenges of Pose Estimation for Shiny Metal Parts}
	\label{fig:challenges}
\end{figure}

The observed appearance of shiny metal parts depends largely on illumination conditions and objects nearby. With slight variation of orientation, the same feature of shiny metal parts in the Fig.\ref{fig:challenge_1a} have drastic color differences. As shown in the Fig.\ref{fig:challenge_1b}, the reflection of a part itself or parts nearby creates additional ambiguities. Furthermore, the partial occlusion of clustered metal parts also should be taken into consideration.

Many pose estimation approaches (e.g. \cite{WangDenseFusion6DObject2019,HodanTLESSRGBDDataset2017}) need the depth information retrieved by active depth-sensing cameras , which are ineffective to shiny metal parts because the projected pattern may have misleading reflection or be undetectable outdoors. Therefore, it is crucial to develop a method that could solely rely on color information. 

Deep learning methods are known to be robust, but their implementations are based on huge amounts of training data, and require intensive manual work for labeling the datasets. To overcome this limitation, it is desirable to have approaches for synthesizing datasets for training \cite{SongExploitingTrademarkDatabases2019, TremblayFallingThingsSynthetic2018, MullerSim4CVPhotoRealisticSimulator2018a, TremblayTrainingDeepNetworks2018} and producing label data \cite{MarionLabelFusionPipelineGenerating2017}\cite{SuchiEasyLabelSemiAutomaticPixelwise2019}.

\subsection{Related Works}
In this paper, we aim at estimating the pose of clustered metal parts. Our approach is based on training with synthetic images. Key concepts and related works in this field are discussed below.

\textbf{Pose Estimation} The study of estimating the 6D pose through a single image is a popular and important topic. The methods generally assume a known rigid object model and a set of corresponding 2D-to-3D point mapping. In this field, many methods have been proposed, e.g. \cite{ZhengRevisitingPnPProblem2013}\cite{FerrazVeryFastSolution2014a}\cite{LepetitEPnPAccurateSolution2009}. This kind of problem is usually called the Perspective-n-Point problem (PnP). A drawback of these approaches is that they fail to address texture-less objects. 
Recent literature about 6D pose estimation using deep learning methods. Hinterstoisser et al. \cite{HinterstoisserModelBasedTraining2013} first employed quantized color gradient and surface normal feature for estimating pose of texture-less objects. After that, Hodan et al. \cite{HodanDetectionfine3D2015} proposed improved template matching method. Zhang et al. \cite{ZhangDetectRGBOptimize2019} estimated the 6D pose for texture-less objects by using edge information. Wohlhart et al. \cite{WohlhartLearningDescriptorsObject2015} employed network to extract features. Kehl etal. \cite{KehlSSD6DMakingRGBbased2017a} presented SSD-6D which relies on the SSD architecture \cite{LiuSSDSingleShot2016} to predict estimated poses. Tekin et al. \cite{TekinRealTimeSeamlessSingle2017} used an architecture to predict some 2D coordinates from which the object pose can be recovered. 
\begin{figure*}[h]
	\centering
	\includegraphics[width=\linewidth]{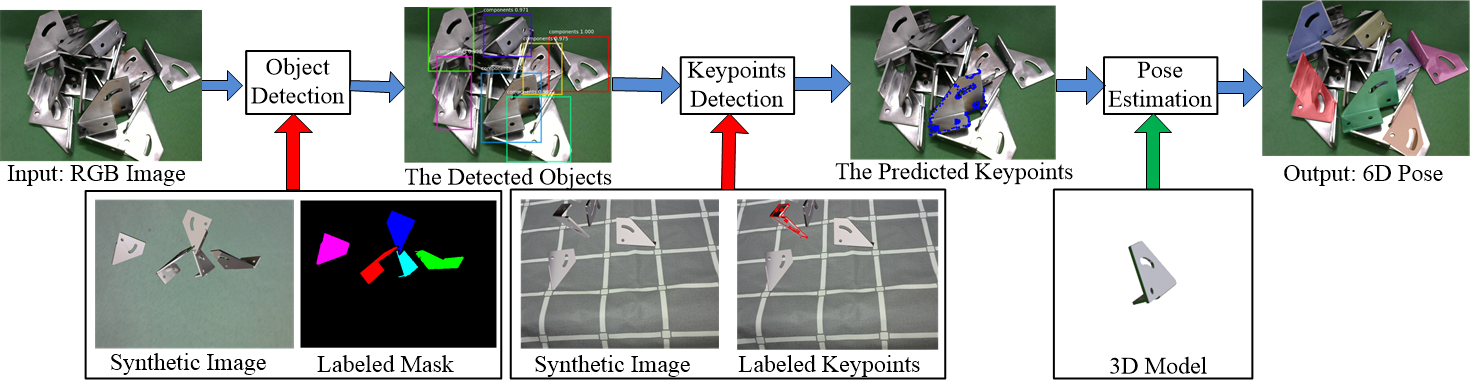}
	\caption{Overview of our approach}
	\label{overview of our approach}
\end{figure*}%

\textbf{Data Synthesis} Current leading object detectors rely on convolutional neural networks. However, to perform at their best, they require huge amounts of labeled training data. Labeling of the data is usually time consuming and expensive. Using synthetic images is therefore very attractive. 
Unfortunately, synthetic rendering methods are usually unable to reproduce the statics produced by their real-world environments. The synthetic images and real images have gaps, as observed in for example \cite{HinterstoisserPreTrainedImageFeatures2017}.
NVIDIA proposed a method \cite{TremblayDeepObjectPose2018} to span the gap through a combination of domain randomized and photorealistic data, this data generation method is the first one trained only on synthesis data that can achieve good performance on pose estimation area. But this method cannot model highly reflective metallic parts well. 

\textbf{Automatic Labeling} Many state-of-the-art pose estimation methods use supervised learning techniques. But annotations in datasets need expensive labor and the accuracy of the ground truth in datasets cannot be guaranteed. The pixel-wise labeling usually is annotated by using an open annotation tool called LabelMe. The EasyLabel \cite{SuchiEasyLabelSemiAutomaticPixelwise2019} is a semi-automatic annotation tool to extract precise object masks. To our knowledge, there is no datasets which mask is automatically annotated. Obtaining the ground truth annotations for object poses is not a trivial task. For most of the recognition tasks, such as object classification or object detection, the ground truth annotations are provided by humans. To benefit the research on object pose estimation, we believe that a large scale dataset with accurately annotated object poses is necessary. On the one hand, supervised learning methods can benefit from more training data. On the other hand, it also poses challenges on the scalability of the existing methods, which can promote new researches on the pose estimation problem. As for keypoints datasets, a well-known one is PASCAL3D+ \cite{XiangPASCALbenchmark3D2014}.They designed a MATLAB tool for annotating the keypoints of objects. In the PASCAL3D annotation process, you can load a CAD model of an object, then you can label the keypoints corresponding to the points annotated in the CAD model. All the work is completed by humans, the precision cannot be guaranteed.

\subsection{Major Contributions}
We have made several contributions to help tackling the task of pose estimation for clustered texture-less shiny objects. 

First, we proposed a novel way of interpreting the detected edges. Specifically, our method does not rely on well estimated continuous contour, but instead, thoroughly investigate different combinations of discrete edges. According to our test, by evaluating reprojection error, the ambiguities caused by reflection and partial occlusion can be eliminated effectively. 

Second, we designed an efficient mechanism to automatically synthesize a large quantity of training data using Blender \cite{FoundationblenderorgHome}. Our method significantly reduced the cost and time consumption for implementing deep learning methods. By altering background and illumination, our dataset could well cover various conditions and therefore guarantee the robustness of the trained neural network. 

Third, we developed an algorithm for automatically labeling the datasets. Compared with manual labeling, our solution not only avoided the influence of uncontrollable human factors (e.g., fatigue and distraction), but also ensured higher accuracy and precision.  According to \cite{SuchiEasyLabelSemiAutomaticPixelwise2019}, the better accuracy of labeling could result in improved performance. 

Forth, we successfully trained multiple neural network for different purposes using the same datasets. In such a way, the demand for training data decreased significantly. Therefore, the system could have lower hardware requirements, especially in size of memory and storage.

\section{SYSTEM OVERVIEW}
Fig. \ref{overview of our approach} provides an overview of our proposed workflow, which mainly comprises three stages. In the first step, the image containing the parts will be put through an object detection neural network. Then, in the detected region of interest (ROI), keypoints are estimated by another deep neural network. Both neural networks are trained with the same dataset we synthesized, which includes rendered images and labeled keypoints. In the end, with the estimated keypoints information, each part is segmented from the image and its 6D pose is derived. 

\emph{\romannum{1})} The convolutional neural network we used for detecting the objects is an off-the-shelf neural network called Mask-RCNN \cite{HeMaskRCNN2017}, which is known to have good performance in object detection tasks. We leveraged Blender \cite{FoundationblenderorgHome} to automatically synthesize all the training data for Mask-RCNN. Labeling the mask in Mask-RCNN needs intensive human labor, so we also proposed a new way to label the mask automatically.

\emph{\romannum{2})} Our keypoint estimation method was inspired by an study that uses keypoints information to predict pose of human \cite{NewellStackedHourglassNetworks2016}. In our system, a convolutional neural network called stacked hourglass network \cite{NewellStackedHourglassNetworks2016} was utilized to predict the keypoints of metal parts. We also automated the keypoint labeling process to avoid overwhelming manual work. For higher accuracy of object pose estimation, we build a new RGB dataset which includes 20K RGB synthesized images. Each image contains multiple metal parts with masks labeled on each metal part and labeled keypoints on each object.

\emph{\romannum{3})} The predicted keypoints on the objects are used to perform segmentation and estimate the 6D pose of each detected object. To tackle pose estimation errors incurred due to erroneously detected keypoints, we use PnP(Perspective-n-Point) combined with Ransac(Random sample consensus) to remove the outliers.

\section{OBJECT DETECTION}
Since it is difficult to segment shiny texture-less objects with traditional way, we use the deep learning network Mask-RCNN to implement it. Mask-RCNN extends the Faster-RCNN \cite{RenFasterRCNNRealTime2017} model for object instance segmentation.We use the ResNet101 as a backbone network to extract features. 
For the purpose of building an object detection network, we use the synthesis data for training. The usage of synthetic data has an advantage that it can avoid overfitting to a particular distribution demonstrated in the dataset. The synthetic data should cover various condition including lighting change, background variation, the reflectance properties. These conditions can be emulated by using the software Blender. The synthetic data are generated using this software by placing the CAD in virtual 3D background scenes with physical constraints. Backgrounds are rendered with real world scenes. The CAD model is placed with various position and orientation. 

After we generate the dataset for object detection, we find a way for pixel-wise labeling automatically. For our dataset,

\begin{figure}
	\centering
	\begin{subfigure}[b]{0.237\textwidth}
		\centering
		\includegraphics[width=\textwidth]{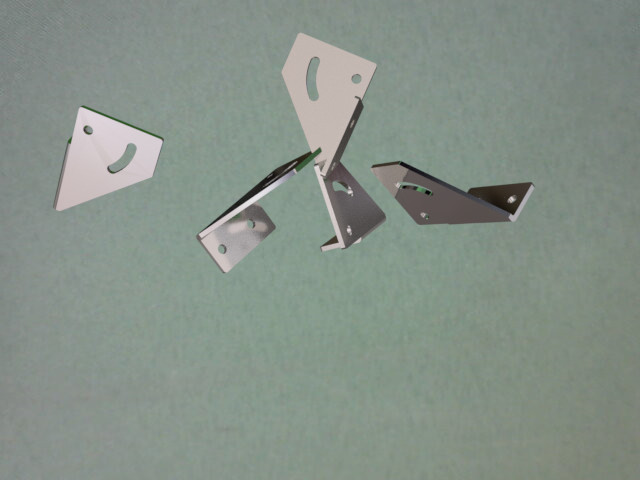}
		\caption{Synthesis images for metal parts}
		\label{fig:the raw one}
	\end{subfigure}
	\hfill
	\begin{subfigure}[b]{0.237\textwidth}
		\centering
		\includegraphics[width=\textwidth]{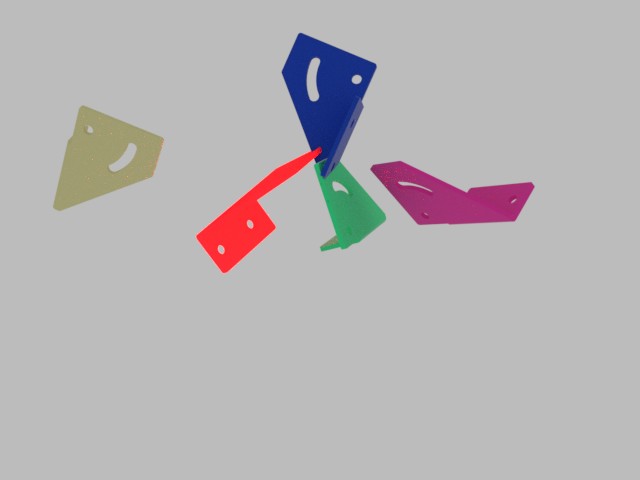}
		\caption{Specified colors for original metal parts}
		\label{specified colors for original metal parts}
	\end{subfigure}
	\hfill
	\begin{subfigure}[b]{0.237\textwidth}
		\centering
		\includegraphics[width=\textwidth]{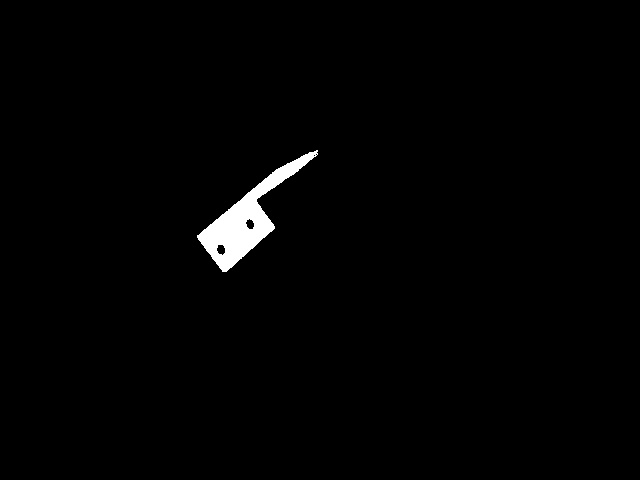}
		\caption{Binary image a part in (b)}
		\label{binary image for a part in color image}
	\end{subfigure}
	\hfill
	\begin{subfigure}[b]{0.237\textwidth}
		\centering
		\includegraphics[width=\textwidth]{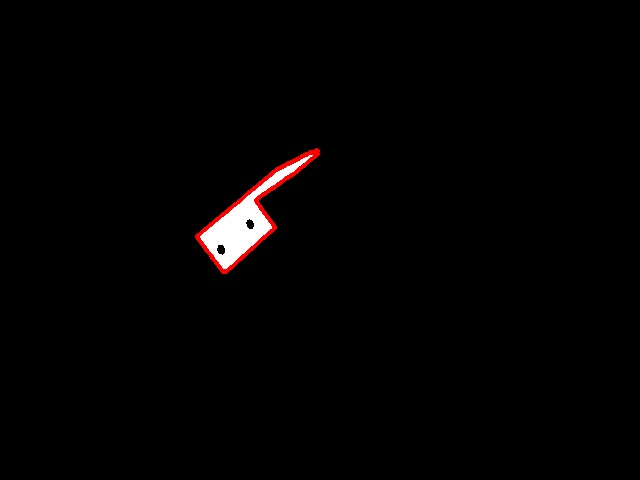}
		\caption{Find the external contour}
		\label{find the max out contour for a part}
	\end{subfigure}
	\hfill
	\begin{subfigure}[b]{0.237\textwidth}
		\centering
		\includegraphics[width=\textwidth]{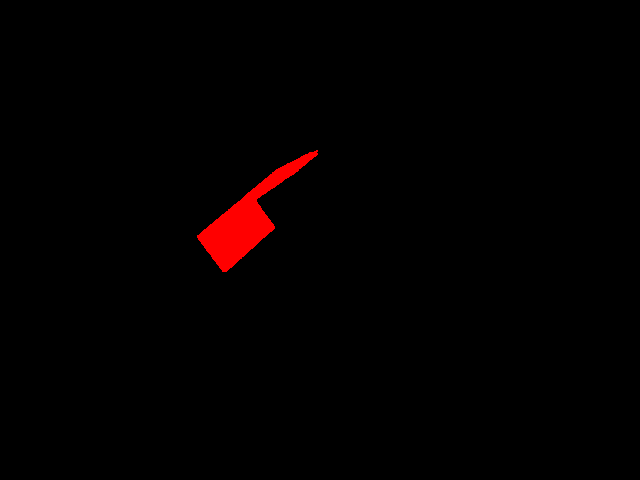}
		\caption{Specified the area of the contour in (d) as mask of the metal}
		\label{the mask image}
	\end{subfigure}
	\hfill
	\begin{subfigure}[b]{0.237\textwidth}
		\centering
		\includegraphics[width=\textwidth]{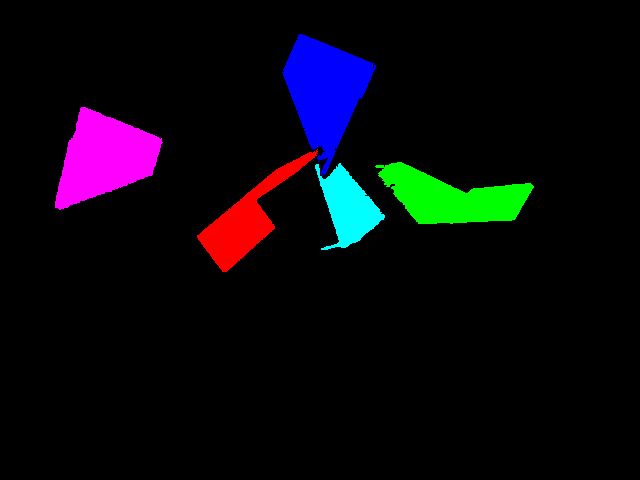}
		\caption{All metal parts are labeled in index image}
		\label{all metal parts are labeled in index image}
	\end{subfigure}
	\hfill
	\caption{The procedure of pixel-wise annotation}
	\label{label mask}
\end{figure}
the main idea involved in automatic mask labeling is that we use Blender to generate images without any background information. And the CAD model are set as non-metallic material like plastics, as shown in the second image in Fig. \ref{label mask}. We binarizate each part individually. Then we find contour of each part and fill its contour, the filled one is the mask, we can achieve the final mask annotation. The challenging aspect within it is to find the contour of each metal part. Because the color of industrial parts looks different with different lighting conditions and positions, it is difficult to detect their contours directly. So the labeling is conducted by using intermediate image as shown in Fig. \ref{label mask}. If the property of a model is set as plastic, then its appearance will be stable with the changing of lighting condition. After we generate the different color CAD model which is reliable to light, we can find the contour for each instance. 
\section{KEYPOINTS DETECTION}
Inspired by the stacked hourglass network architecture used in human pose estimation \cite{NewellStackedHourglassNetworks2016}, we use the network as a detector to detect the semantic keypoints on the edge. This network usually contains two hourglass modules. 

The module is from high to low resolution and then from low to high resolution. The local and global features information can be obtained and combined to determine the location of the landmarks. The 2D object keypoint locations are represented by the heatmaps. 

It is difficult to label dense semantic keypoints precisely by humans, so we propose to automatically label the keypoints in synthesis data. The first thing is to generate the bounding box around the metal parts so that we can detect the metal parts well. 
 
When we generate the synthesis image, we can produce the corresponding metal part alone without any background in the PNG image format. We separate the PNG image into four channels: RGB channel and alpha. The alpha channel is used to judge whether the pixel belongs to the object.

The dataset contains kinds of images in different scenarios. We use the images taken from the real world as the background, then we can generate kinds of synthetic images with different backgrounds. Through this way, we can acquire different kinds of training data with annotated keypoints. The intrinsic camera parameters of Blender virtual camera can be set. Taking an industrial parts for example, this process is composed of sequential steps of loading CAD models, choosing landmarks and labeling keypoints.

\begin{figure}
	\centering
	\includegraphics[width=\linewidth]{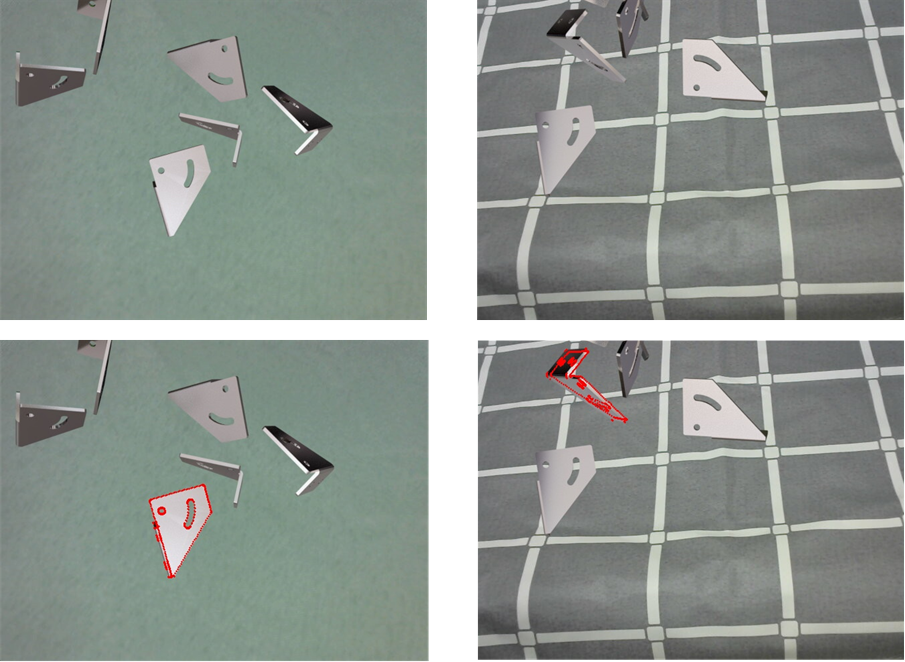}
	\caption{The result of label keypoints}
	\label{The result of label keypoints}
\end{figure}

The keypoints labeling result is shown in Fig. \ref{The result of label keypoints}.
\begin{figure}
	\begin{subfigure}[b]{0.23\textwidth}
		\centering
		\includegraphics[width=\textwidth]{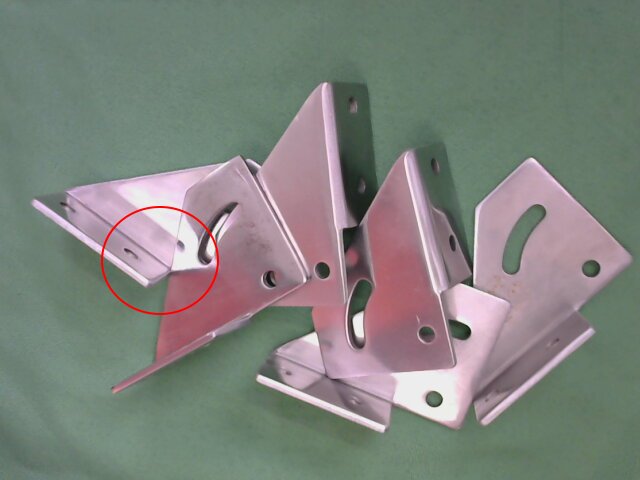}
		\caption{}
		\label{1}
	\end{subfigure}
	\hfill
	\begin{subfigure}[b]{0.23\textwidth}
		\centering
		\includegraphics[width=\textwidth]{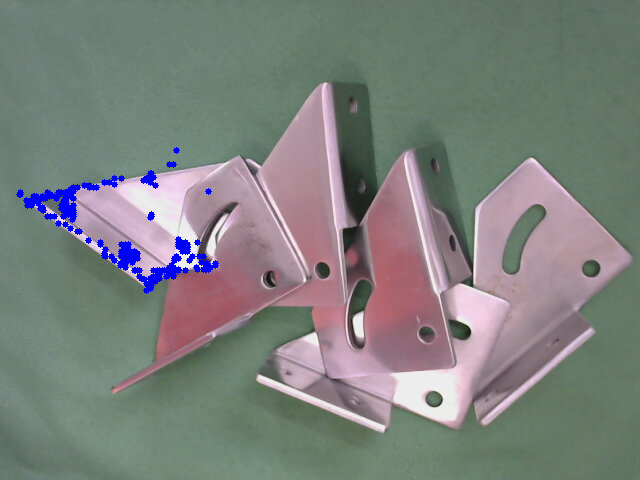}
		\caption{}
		\label{2}
	\end{subfigure}
	\hfill
	\begin{subfigure}[b]{0.23\textwidth}
		\centering
		\includegraphics[width=\textwidth]{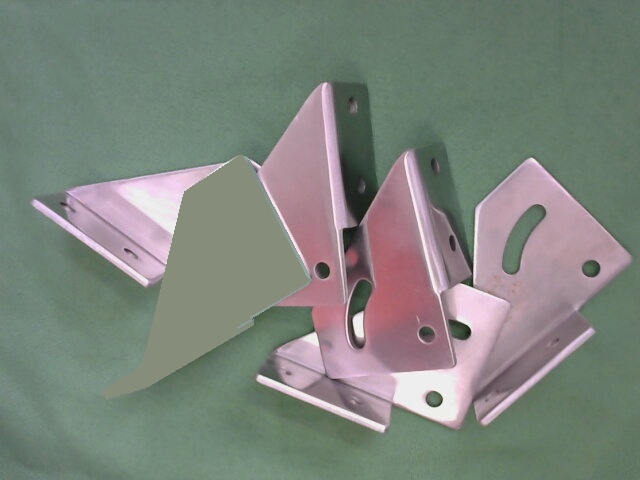}
		\caption{}
		\label{3}
	\end{subfigure}
	\hfill
	\begin{subfigure}[b]{0.23\textwidth}
		\centering
		\includegraphics[width=\textwidth]{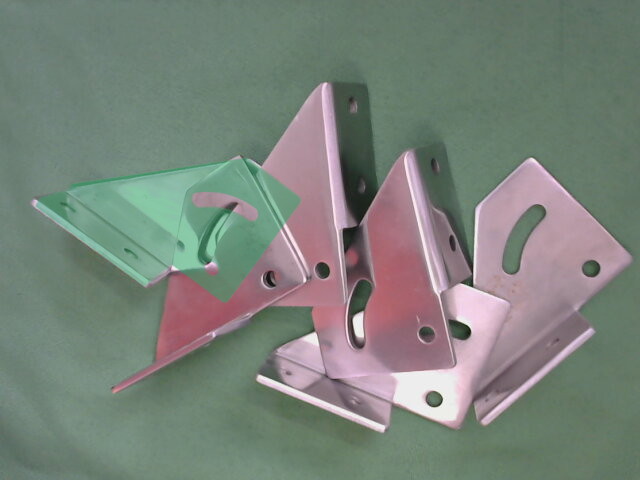}
		\caption{}
		\label{4}
	\end{subfigure}
	\caption{Pose optimization}
	\label{pose_optimization}
\end{figure}

\section{POSE OPTIMIZATION}
As for texture-less metal parts, their CAD models can be obtained easily from mechanical design software. Given the CAD model, we can obtain the 3D keypoints and their corresponding keypoints projected in 2D. When we use the predicted keypoints directly, the error in keypoint prediction will affect the results. To lessen the effect of the uncertainty in 2D keypints predictions, a parameter $d_{i}$ is used to indicate the localization confidence of the keypoint in the image. $d_{i}$ represents the peak value in the heatmap mapping to the \emph{i}th keypoint. The 3D model is known, the keypoint locations on the 3D model are known either. The corresponding 2D positions of the keypoints in the images can be obtained after using the stacked hourglass network module. Some predicted keypoints are inaccurate due to some false detections. To address this problem, we propose to use the PnP and RANSAC to remove the outliers, and then use the CAD model to combined with the weights to each keypoint to refine the pose.

Since the metal parts are clustered together, we should consider the influence made by inter-reflection. This problem can be well demonstrated in the example in Fig. \ref{1} where in the region marked by red circle two overlapped parts reflect to each other. And the left part is partially reflected on the right part. A direct keypoints estimation applied to the region will produce the result as shown in Fig \ref{2}, where one portion of the right parts is recognized as the left part. Our solution for this problem is firstly to estimate the pose of the right part. Then with the estimated pose, the projection of the part in 2D plane is re-rendered by the color of the background. This can be achieved by using OpenDR \cite{LoperOpenDRApproximateDifferentiable2014} as shown in Fig. \ref{3}. Then keypoint estimation applied to the revised image without the right part will provide the accurate position estimation as shown in Fig. \ref{4}.

\begin{figure}
	\centering
	\includegraphics[width=\linewidth]{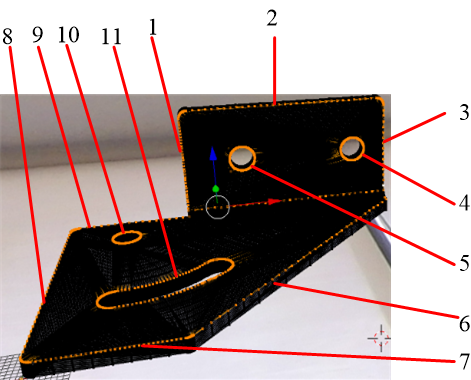}
	\caption{The number sequence of the metal parts edge}
	\label{number metal}
\end{figure}

When the edge and the neighbor area of metal are shiny, it is difficult to use only edge information to estimate the pose. We number the metal edge as shown in Fig. \ref{number metal}. If some edge is shiny, we can use other edges for pose estimation. 
\begin{figure*}
	\centering
	\includegraphics[width=\textwidth]{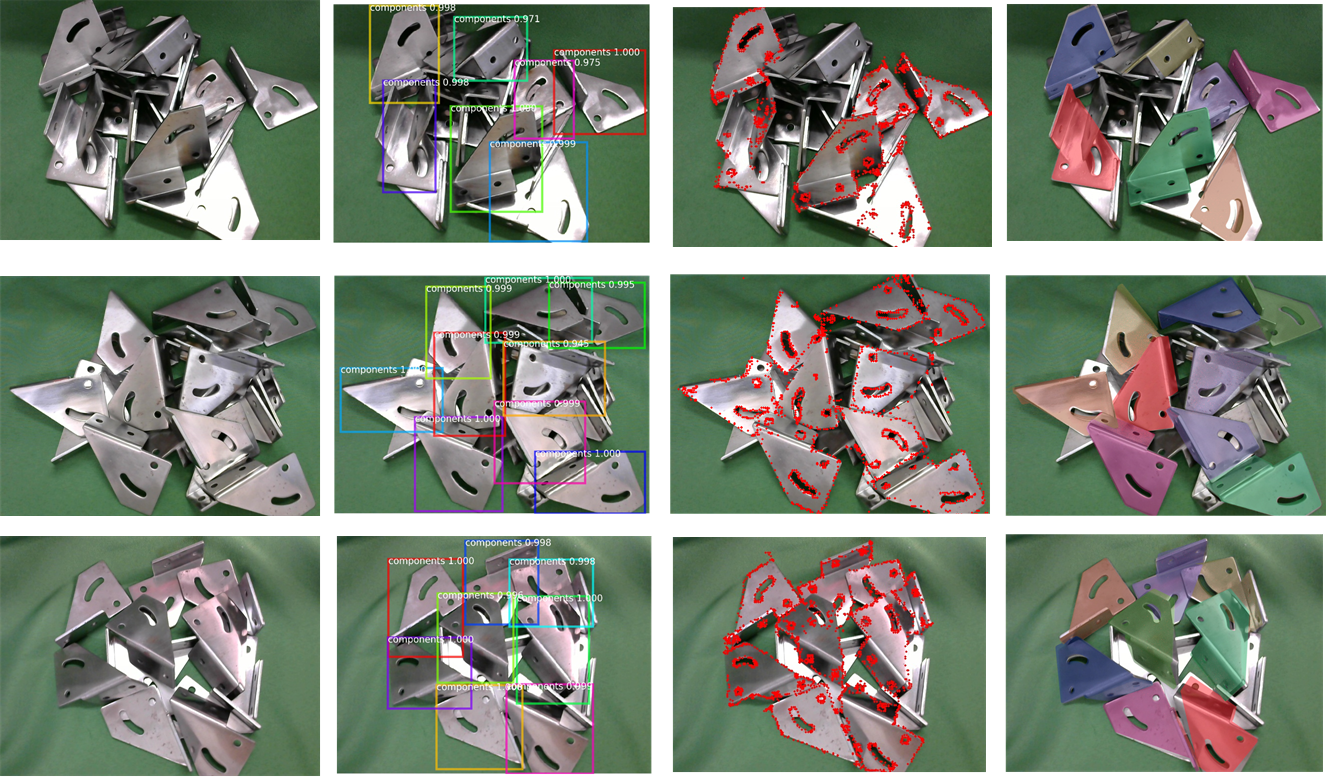}
	\caption{The results of pose estimation: the images in the first column are the raw images taken by Logitech camera, the images in the second column are the detected metals parts with bounding box, the images in the third column are predicted keypoints images marked with red dots, the images in the last column are the projections of 3D metal parts model with estimated poses}
	\label{The results of pose estimation}
\end{figure*}
\section{Experiments}
In this section, we use a Logitech C310 HD camera to estimate metal parts pose and render the object to the original image based on the estimated pose to show the effect of pose estimation.
\subsection{Object Detection}
\begin{figure}
	\centering
	\includegraphics[width=3.5in]{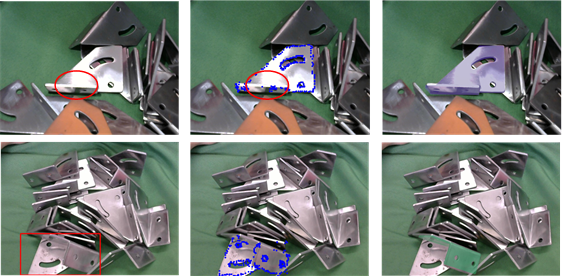}
	\caption{The example of pose estimation for metal parts}
	\label{shiny edge}
\end{figure}
In the object detection part, we use Blender to produce about 20K images, each image contains about 5 or 6 metal parts. All the images are with different backgrounds. The Mask-RCNN \cite{HeMaskRCNN2017} is used as the detection network. We use synthetic images and labeled annotation as the training data. The detection results are shown in the second column in  Fig. \ref{The results of pose estimation}.

\subsection{Keypoints detection}
We choose about 700 keypoints on the CAD model along the edge and project the 3D keypoints to the 2D images for training and testing in the stacked hourglass network \cite{NewellStackedHourglassNetworks2016}. The chosen keypoints are described in Fig. \ref{number metal} with orange dots as input 3D model. In this part, we use about 4000 metal parts images with different rotational and translational positions. A random 90\%10\% split is used for keypoints training and test data. The result is shown in Fig. \ref{The results of pose estimation}. In our pose estimation procedure, we use OpenDR\cite{LoperOpenDRApproximateDifferentiable2014} to render the object which has been applied for pose estimation before to reduce the influence made by inter reflection by each other. The sequence of predicting metal parts are decided by the confidence of object detection. The better confidence indicates that the corresponding part is recognized better. Then it is reasonable to apply pose estimation to the part preferentially.

\subsection{Pose estimation}
As shown in the first row in Fig. \ref{shiny edge}, when some edges are shiny, we can choose the other edges as the features to obtain the pose. When objects are partially occluded by others, as shown in the second row, the keypoints will be scattered on different metal parts. In this case, the obtained keypoints will be clustered by different pose candidates. The reprojection error is used to choose which result is the most reasonable one. The results following this strategy are shown in the last column in Fig. \ref{The results of pose estimation}. 

\subsection{Processing time}
On a desktop with an Intel i7 3.4GHz CPU, 8G RAM and a GeForce GTX Titan Xp 12GB GPU, our pipeline needs around 1.2 seconds for object detection and 0.5 seconds for keypoints detection in one instance and less than 0.3 seconds for pose optimization, the total running time depends on the number of metal parts detected.

\section{CONCLUSIONS}
In this project, we proposed a new approach for estimating the pose of clustered texture-less shiny objects. We developed a new mechanism for automatically synthesizing and labeling our own dataset, which has been utilized to train two different neural networks in our system. In our tests, the trained neural networks well accomplished the task of object detection and key point estimation. With these results, we successfully estimated the pose of target objects in real environment and therefore validated the effectiveness of our method. Our solution expanded the genre of objects industrial robot can handle and thus have great value for the evolution of industrial automation.

\addtolength{\textheight}{-3cm}   






\bibliographystyle{IEEEtran}
\bibliography{IEEEabrv,IEEEexample,mylibrary}

\end{document}